\documentclass{WileyMSP-template}

\usepackage{xurl}
\usepackage{booktabs}
\usepackage{nicefrac}
\usepackage{multirow}
\usepackage{ulem}
\usepackage{siunitx}
\usepackage[utf8]{inputenc} % allow utf-8 input
\usepackage[T1]{fontenc}    % use 8-bit T1 fonts
\usepackage{hyperref}
\hypersetup{
    colorlinks=true,
    draft=false,
    urlcolor=blue,
    linkcolor=red,
    citecolor=blue
}
\usepackage[font=footnotesize,labelfont=bf ]{caption}
\usepackage{bbm}
\usepackage{bm}
\usepackage{amsmath,amssymb,amsfonts,amsmath}
\usepackage[usenames,dvipsnames,svgnames,table]{xcolor}
\usepackage[capitalise]{cleveref}

\usepackage{wrapfig}
\usepackage{float}
\usepackage{dsfont}
\usepackage{graphicx}
\usepackage{microtype}
\usepackage{setspace}
\usepackage{soul}
\usepackage{xcolor}
\usepackage{sectsty}
\usepackage[numbers,sort&compress]{natbib}

\subsubsectionfont{\fontsize{10}{15}\selectfont}
\sectionfont{\LARGE}
\subsectionfont{\Large}

\usepackage[indent=10pt]{parskip}

\usepackage[fontsize=9pt]{fontsize}

\begin{document}
\pagestyle{fancy}
\hspace{-19.5pt}\title{\textbf{Learning, locomotion, and navigation of soft synthetic snakes in three-dimensional, heterogeneous environments}}
%\hspace{-19.5pt}\title{\textbf{Navigation of soft limbless robots in three-dimensional, heterogeneous environments}}
\maketitle

% Author: Please give full first and last names for authors and include * after the name of all corresponding authors
\hspace{-18pt}\author{Xiaotian Zhang,}
\author{Ali Albazroun,}
\author{Tixian Wang,}
\author{Songyuan Cui,}
\author{Prashant G. Mehta,}
\author{Mattia Gazzola\textsuperscript{*}}

% Affiliations: Please provide academic titles (Prof. or Dr.) for all authors where applicable, and include an institutional email address for all corresponding authors
\begin{affiliations}
\hspace{-13pt}Xiaotian Zhang, and Mattia Gazzola\\
Carl R. Woese Institute for Genomic Biology, University of Illinois Urbana--Champaign\\

\vspace{10pt}
Xiaotian Zhang\\
Department of Mechanical and Aerospace Engineering, Hong Kong University of Science and Technology\\

\vspace{10pt}
Ali Albazroun, Tixian Wang, Songyuan Cui, Prashant G. Mehta, and Mattia Gazzola\\
Department of Mechanical Science and Engineering, University of Illinois Urbana--Champaign\\
Email Address: mgazzola@illinois.edu

\end{affiliations}

\vspace{10pt}
%\noindent\normalsize{\textbf{Keywords: }{Open-source system; Electrophysiology; In vitro neural interfaces; Neural computing}}

\justifying

\begin{abstract}
\noindent\normalsize{\textbf{Abstract.}}
Limbless terrestrial animals exhibit exceptional locomotor versatility and control, currently unmatched by engineered counterparts. Here, we introduce a computational framework that enables soft synthetic snakes to navigate unstructured, heterogeneous 3D terrains. Our approach is grounded in bio-inspired actuation and sensing models that reduce the control complexity inherent to high-degree-of-freedom, continuum bodies. These models are integrated into a reinforcement learning architecture to derive environment-traversing policies. Training first occurs in simplified, homogeneous terrains to learn locomotion primitives. These are then composed into adaptive strategies for complex landscapes. We demonstrate robustness by deploying a snake in high-fidelity 3D environments reconstructed from real-world imaging, achieving reliable navigation. Overall, this work provides a physically-realistic simulation platform and practical insights for the control of continuum systems in natural terrains.
%Soft limbless animals exhibit extraordinary locomotor versatility in natural environments, yet achieving comparable performance in engineered counterparts remains a fundamental challenge. Here, we present a computational framework that enables soft synthetic snakes to navigate complex, three-dimensional, and heterogeneous terrains. Our approach is grounded in bio-inspired actuation and sensing models that effectively reduce the control complexity inherent to high-dimensional, continuum snake bodies. These models are then embedded within a reinforcement learning framework to derive navigation policies. First trained in simplified, homogeneous environments, robust locomotion strategies are obtained, forming the basis of a modular decision-making mechanism that enables adaptive gait modulation across increasingly complex terrains. The robustness of this framework is ultimately demonstrated by deploying the snake in high-fidelity 3D environments reconstructed from real-world imaging data, where successful navigation is achieved despite substantial environmental variability. Overall, this work provides an instrumental simulation platform and practical control insights for the deployment of soft slithering systems in real-world terrains.
\end{abstract}

\section{Introduction}

Despite relatively simple body plans, limbless terrestrial creatures achieve striking locomotor performance by exploiting continuum mechanics and distributed substrate contact~\cite{Mosauer:1932,Gray:1946,Goldman:2010}. Snakes exemplify this versatility, traversing complex terrain via slithering, sidewinding, climbing, jumping, and even aerial gliding~\cite{Gray:1950,Jayne:1988,Marvi:2013,Socha:2002,Jayne:2020,Graham:2023,Tingle:2024}. Inspired by these systems, recent years have seen growing engineering interests toward slender robots capable of navigating unstructured, heterogeneous, and confined settings, for defense, exploration, inspection, and medicine~\cite{Hirose:1993,Tanev:2005,Onal:2013,Qi:2020,Seetohul:2022,Yasa:2023}.
%Limbless terrestrial creatures, despite possessing relatively simple body plans, achieve remarkable locomotor performance by exploiting continuum mechanics and distributed contact~\cite{Mosauer:1932,Gray:1946,Goldman:2010}. These traits allow snakes, for example, to excel at traversing complex landscapes, employing a diverse behavioral repertoire including slithering, sidewinding, climbing, jumping, and even flying~\cite{Gray:1950,Jayne:1988,Marvi:2013, Socha:2002,Jayne:2020,Graham:2023,Tingle:2024}. Long studied in biology, limbless locomotion has attracted growing engineering interest over the past 15 years, in pursuit of robotic systems able to navigate unstructured, heterogeneous, and confined environments, for applications in defense, maintenance, or medicine~\cite{Hirose:1993,Tanev:2005,Onal:2013,Qi:2020,Seetohul:2022,Yasa:2023}.

However, realizing this potential in artificial systems remains challenging. Together with materials and fabrication difficulties, control is a primary bottleneck. Indeed, soft slender bodies exhibit high-degree-of-freedom (high-DOF), strong nonlinearity and contact-rich terrain interactions that defeat tractable control. As a result, model-based methods remain limited in scope and capability~\cite{Santina:2023}. Model-free methods, especially those leveraging unsupervised learning, offer an alternative by deriving controllers directly from experience~\cite{Jia:2021,Bing:2022,Yasa:2023,Zhang:2025}. Yet, their dependence on large datasets and intensive computation hinders scalability and physical deployment~\cite{Mengaldo:2022}. Consequently, existing robotic prototypes remain constrained to simplified settings~\cite{Transeth:2008,Jia:2021,Mengaldo:2022,Zhang:2025} with limited real-world applications.

%However, realizing this potential in artificial systems remains challenging. Beside material fabrication, control is a significant limiting factor. Indeed, the nonlinear, elastic, and continuum nature of soft, slender bodies often introduces high-dimensional, non-linear dynamics that, when coupled with complex terrain interactions, render control hard. Model-based approaches often fail in such settings, due to difficulties in capturing structural, material, and environmental effects~\cite{Santina:2023}. Model-free methods, especially those leveraging Reinforcement Learning (RL), offer an alternative path by deriving control laws directly from experience~\cite{Jia:2021,Bing:2022,Yasa:2023,Zhang:2025}. Yet, their reliance on large datasets and/or heavy computations impairs scalability and hinders physical deployment~\cite{Mengaldo:2022}. As a result, existing robotic prototypes remain constrained to simplified settings~\cite{Transeth:2008,Jia:2021,Mengaldo:2022,Zhang:2025}, limiting real-world application.

In this context, we present a hybrid approach that embeds a continuum-based numerical model within a reinforcement learning (RL) framework, demonstrated through the control of soft synthetic snakes traversing three-dimensional, heterogeneous environments.
We develop a computational environment that captures the coupled dynamics between a continuum elastic filament (snake) and high-fidelity terrains reconstructed from real-world imaging data. Building on this, we propose an RL-based control strategy informed by bio-inspired models of actuation and sensing. Our approach leverages the concept of \textit{physical intelligence}~\cite{Ulrich:1988,Pfeifer:2007,Mengaldo:2022,Wang:2023}, whereby passive body mechanics and terrain-mediated interactions are deliberately exploited to relieve neuromuscular control.
Consistent with this principle, we employ a compact library of stereotyped actuation templates (motor primitives)~\cite{Marvi:2014,Astley:2015,Zhang:2021} to simplify the coordination of high-DOF body dynamics.
In limbless locomotion, such templates have been shown to generate a diverse repertoire of canonical gaits~\cite{Astley:2015,Zhang:2021} and to enable traversal across varying---albeit simplified---frictional environments via passive adaptation~\cite{Zhang:2021}.

%In this context, we present a hybrid approach that integrates a continuum-based numerical model into a reinforcement learning (RL) framework, showcased via the control of a soft synthetic snake navigating three-dimensional, heterogeneous environments. To this end, a computational framework is developed to simulate the dynamic interaction between a continuum elastic filament (the snake) and high-fidelity terrains reconstructed from real-world imaging data. Building on this foundation, we propose a RL-based control strategy grounded in bio-inspired models of actuation and sensing. The strategy leverages the concept of ``physical intelligence''~\cite{Ulrich:1988,Pfeifer:2007,Mengaldo:2022,Wang:2023}, wherein passive mechanics and environmental interactions are purposefully exploited to relieve neuromuscular control. Embodying this principle, a small set of stereotyped actuation templates (or motor primitives)~\cite{Marvi:2014,Astley:2015,Zhang:2021} can be employed to simplify the coordination of high-DOF body dynamics. In limbless locomotion, templates have been shown to effectively produce a variety of stereotypical gaits~\cite{Astley:2015,Zhang:2021}, while enabling traversal across varying (albeit simplified) frictional environments through passive adaptation~\cite{Zhang:2021}.

We further integrate a bio-inspired sensory feedback strategy, broadening the controller's effectiveness on complex, unstructured terrains. Our sensing models abstract biological neural circuits as coupled populations of oscillators~\cite{Tilton:2014, Wang:2019, Wang:2020, Naughton:2026}, enabling robust estimation of both internal states (body configuration) and external states (environmental features). These sensory signals close the loop to adapt actuation, and are integrated within the RL framework to learn effective terrain-traversal and navigation policies. Equipped with this architecture, synthetic snakes reliably reach prescribed target locations across increasingly challenging topographies, demonstrating the robustness and versatility of the proposed modeling and control approach.

%This concept is compounded here with sensory feedback, greatly expanding the control strategy's versatility and effectiveness in the face of complex, unstructured terrain features. Specifically, we employ sensing models that abstract biological neural networks as populations of oscillators~\cite{Tilton:2014, Wang:2019, Wang:2020, Naughton:2026}, enabling robust estimation of both internal (body configuration) and external (environmental) states. The sensory signals are in turn used to close the loop and adapt actuation, which are eventually integrated within an RL framework to learn effective terrain-traversing and navigation strategies. A synthetic snake equipped with this machinery is demonstrated to successfully reach prescribed target locations across increasingly challenging landscape topographies, showcasing the robustness and versatility of the proposed modeling and control strategy.

Overall, this work integrates templated actuation, robust sensing, modular feedback control, and learning to advance the real-world deployment of soft, limbless robots.

\section{Modeling and control}\label{snake_model}

Here we provide an overview of our modeling, simulation and control framework.
We begin by describing the computational models that describe the snake's body mechanics, its environments, muscular actuation, and sensory capabilities.
We then integrate these physics-based models with RL to derive lightweight, hybridized, and biologically-informed control strategies (Figure~\ref{Fig:Framework_Overview}a).

% \suggest{
% We outline in this section the analytical and computational framework employed to model, simulate, and control the body mechanics of a snake, while fostering multi-modal interactions with complex terrains.
% We begin by introducing the continuum theory leveraged to model the snake's intrinsic mechanics, followed by a description of the computational model that captures its dynamical motion (\cref{section_snake_mechanics}), muscular actuation (\cref{section_muscular_activation}), and sensory capabilities (\cref{section_oscillators}).
% The framework is then extended to incorporate realistic, complex domains based on imaging data sourced externally.
% Finally, we outfit the physics-based model with RL to derive a lightweight, hybridized, biologically-informed control strategy (\cref{section_RL_implementation}).
% }

\subsection{Dynamical modeling of snakes}\label{section_snake_mechanics}

Leveraging the intrinsic slenderness and compliance of the snake body, we model it via the Cosserat rod theory, whereby three-dimensional dynamics and deformations (stretch, bend, twist, and shear) are described in a Lagrangian, one-dimensional fashion~\cite{Antman:1973, Gazzola:2018}.
This representation is particularly well-suited for our study, as it naturally captures the mechanical interplay between snake's body, internal actuation, and environment~\cite{Gazzola:2018, Zhang:2019, Zhang:2021, Tekinalp:2024, Naughton:2021}, while reducing computational costs.
As illustrated in Figure~\ref{Fig:Framework_Overview}b, the rod is kinematically described by a centerline $\mathbf{x}(s, t) \in \mathbb{R}^3$ and a local frame of reference $\mathbf{Q}(s, t)\in \mathbb{R}^{3\times3}$ parameterized by $s \in [0, L]$, where $L$ denotes the rod's length and $t$ is the time.
Rod dynamics are captured via linear and momentum balance, as described in Section~\ref{section_method_CosseratRod}.
Additional forces $\mathbf{F}$ and torques $\mathbf{C}$ are incorporated along the rod, allowing the modeling of distributed muscular actuation, environmental contact, friction, and gravity.
The contact model employed here~\cite{Gazzola:2018, Zhang:2019, Zhang:2021} includes ground reaction forces $\mathbf{n}_g$ and anisotropic Coulomb friction forces $\mathbf{f} = \boldsymbol{\mu} \mathbf{n}_g$, where $\boldsymbol{\mu} = \{\mu_f, \mu_t, \mu_b\}$ denotes the friction coefficients in the forward, transverse, and backward directions, respectively.
In limbless terrestrial locomotion, the anisotropy ratio $\mu_t / \mu_f$ serves as the key parameter that characterizes substrate interactions, while the effect of $\mu_b$ is typically negligible~\cite{Alben:2013, Zhang:2021}.
Further details about our Cosserat rod formulation are provided in Section~\ref{section_method_CosseratRod}.

To numerically implement the rod and contact model described above, we use \textit{Elastica}~\cite{Gazzola:2018, Zhang:2019}, an open-source software for Cosserat rod simulations demonstrated across a range of engineering and biophysical applications, from the design and control of soft~\cite{Naughton:2021} and biohybrid~\cite{Pagan-Diaz:2018, Aydin:2019, Wang:2021, Kim:2023} robots, to the modeling of biological structures such as pigeon wings~\cite{Zhang:2019}, octopus arms~\cite{Shih:2023, Tekinalp:2024}, and snakes~\cite{Zhang:2019, Zhang:2021}.
In this study, we instantiate a rod matching the geometrical and biomechanical properties of the corn snake (\textit{Pantherophis guttatus}~\cite{Gimmel:2021}), a popular animal model~\cite{Fu:2022, Schiebel:2023}.
Further details are provided in Section~\ref{section_method_CosseratRod} and the Supplementary Information (SI).

\subsection{Realistic terrains and contact detection}\label{section_realistic_terrains}
An important departure of this work from previous studies~\cite{Gazzola:2018,Zhang:2019,Zhang:2021} is the inclusion of realistic terrains.
This is achieved by equipping \textit{Elastica} with the ability to import arbitrary 3D geometries into the simulation environment, via the OBJ (Wavefront) file format.
This widespread format efficiently encodes surfaces as collections of vertices, faces, and outward normals, offering a scalable option for representing detailed, complex environmental features.
Importantly, it supports models derived from imaging and scanning data obtained via a variety of sources (e.g., photogrammetry, CAD designs), enabling the study of snake locomotion in high-fidelity, real-world terrains.
Figure~\ref{Fig:Framework_Overview}a illustrates one such example, where the snake is deployed on a Martian landscape model acquired by the Mars Perseverance Rover~\cite{Bell-III:2022}.

The use of meshed 3D landscapes can introduce significant computational costs, particularly in detecting rod-terrain contacts (Figure~\ref{Fig:Framework_Overview}b) when high-resolution topographies are considered.
To address this, we partition and index the terrain into a structured grid of smaller domains, each containing a subset of faces.
At each simulation time step, contact detection is restricted to localized regions surrounding the snake. This strategy markedly reduces computational overhead while maintaining accurate modeling of contact dynamics.
Details can be found in the SI.

\subsection{Muscular actuation}\label{section_muscular_activation}

\begin{figure}[!tb]
\centering
\includegraphics[width=0.99\textwidth]{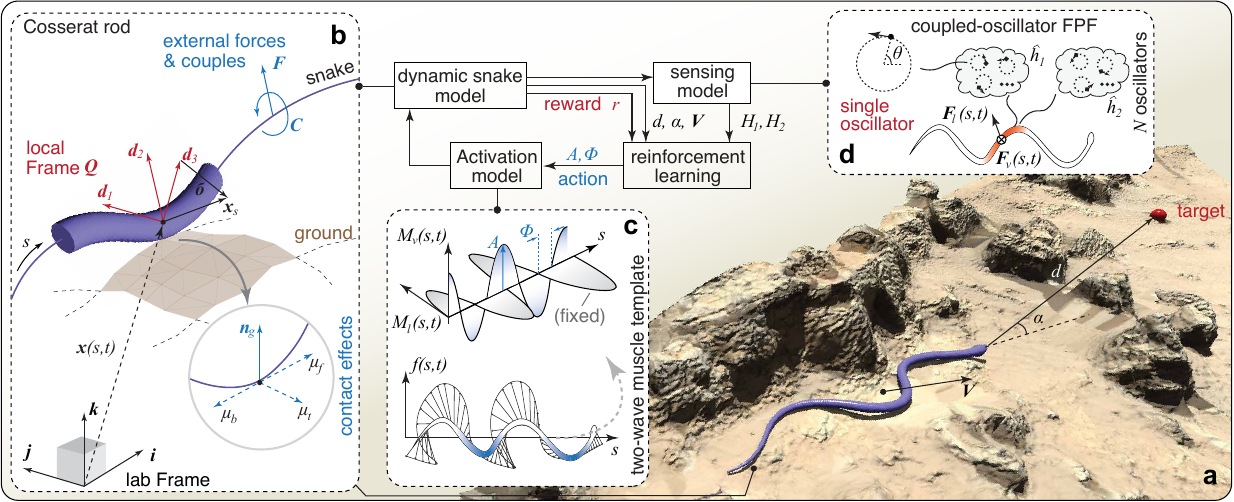}
\caption{
    \textbf{Control framework overview.}
    \textbf{(a)} Illustration of snake navigating 3D complex terrain through implementing a reinforcement learning framework that integrates a dynamic snake model, muscular actuation and sensory feedback.
    \textbf{(b)} Schematic of Cosserat rod representation of snake dynamics.
    Each snake segment is subject to external forces and couples, which models the effects of ground supporting forces, frictional forces and muscular activations.
    Details about the simulation setup and the computational approach for resolving contacts between the snake body and 3D, mesh-based ground can be found in the \Cref{section_method_CosseratRod} and SI.
    \textbf{(c)} Schematic of the two-wave muscular activation template, as well as its effect of modulating underlying frictional forces along the snake body.
    \textbf{(d)} Implementation of coupled-oscillator feedback particle filters to form the state estimation of the snake in the RL framework.
    Two oscillators groups are implemented for sensing lateral ($\bar{h}_1$) and vertical ($\bar{h}_2$) ground forces of the middle portion of the snake.
}\label{Fig:Framework_Overview}
\end{figure}

Based on biological, robotic, and modeling insights~\cite{Hu:2009, Astley:2015}, we implement a two-wave actuation template to simplify the coordination of the snake's high-DOF body dynamics.
This consists of two biologically-observed traveling torque waves (head to tail), one in the vertical ($\mathbf{M}_v$) direction and one in the lateral ($\mathbf{M}_l$) direction.
A simple yet effective control mechanism emerges by fixing the lateral wave as the baseline undulation while actively modulating the vertical wave to lift specific portions of the body~\cite{Zhang:2021}.
As illustrated in Figure~\ref{Fig:Framework_Overview}c, lifting modulates the body contact regions and redistributes the normal forces exerted on the terrain.
This, in turn, alters frictional dynamics, since friction is eliminated (or greatly reduced) where the body lifts, and is enhanced at the remaining contact points where weight is concentrated.
Such mechanism produces net force and torque imbalances over an undulation cycle, enabling directed locomotion and turning maneuvers~\cite{Zhang:2021}.

We thus adopt the framework introduced in~\cite{Zhang:2021}, in which muscular activations in both lateral and vertical directions are prescribed as sinusoidal traveling waves $M(s,t) = A(t)\beta(s)\text{sin}[2\pi (t/T-\Phi -s)]$, where $T$ denotes the undulation period and $\beta(s)$ is a cubic B-spline function defining the torque envelope for each wave.
In this study, both $T$ and $\beta$ are held constant, while the non-dimensional activation ratio $A$ and phase offset $\Phi$ (both $\in [0, 1]$) serve as tunable parameters controlling amplitude and timing of the waveforms.
As discussed previously, we fix lateral undulations (i.e. $\{A_l, \Phi_l\} = \{1, 0\}$) to reproduce lateral curvatures consistent with biological observations~\cite{Hu:2009}, leaving only the two free parameters $A_v$ and $\Phi_v$ of the vertical wave (hereafter referred to as $A$ and $\Phi$).
This leads to a compact action space for RL implementations, highlighting the advantage of incorporating a biologically-informed model.
Full details about the activation parameters are provided in the SI.

\subsection{Neuron-inspired sensing model}\label{section_oscillators}
In addition to intrinsic actuation, robust locomotion requires the snake to sense and respond to environmental features in real time.
Here, we implement a biologically inspired sensory framework based on the coupled oscillator Feedback Particle Filter (FPF) method~\cite{Yang:2013,Tilton:2014,Taghvaei:2014,Wang:2020} (Figure~\ref{Fig:Framework_Overview}d).
FPF exploits the collective dynamics of an ensemble of coupled mathematical particles (oscillators) to provide robust state estimates in the presence of noise, for periodic systems.  We then employ this methodology to sense local frictional environmental characteristics.

To implement this sensing strategy, we deploy two independent set of coupled oscillators along the midsection of the snake body, tasked with sensing ground contact forces in the lateral ($F_l$) and vertical ($F_v$) directions, respectively.
Each group consists of $N$ oscillators, with each oscillator defined by a virtual phase variable $\theta^i_j$ ($i = 1, \dots, N$ and $j = 1, 2$) that evolves periodically within $[0, 2\pi]$.
The dynamics of each oscillator is governed by a combination of intrinsic oscillatory behavior, intra-group coupling, and external noisy measurements (contact forces).
This architecture is reminiscent of biological neural networks, in which neurons' firing emerges from the interplay of spontaneous activity, synaptic interactions, and external stimuli.
The full formulation is provided in Section~\ref{section_method_oscillators}.

The feedback from each oscillator group is captured through the observation functions $h_j$, analogous to a biological perceptual process.
Numerically, we follow the approach of~\cite{Wang:2020} and construct $h_j$ as a linear combination of Fourier basis functions of $\theta^i_j$, with the associated weights updated online during simulation (see Section~\ref{section_method_oscillators}).
This formulation allows each group to adaptively encode the sensed signal as the snake moves, with intra-group coupling promoting oscillators' synchronization.
Upon synchronization, the individual $h_j$ values converge toward their group mean $\hat{h}_j$, which provides an accurate estimate of the corresponding noisy force signal (in our case, $F_l$ and $F_v$).
%Here, $F_l$ encodes information about the ground's frictional anisotropy, while $F_v$ reflects the snake's contact dynamics with the substrate.
%Both signals are inherently noisy (Figure~\ref{Fig:Result_Heter}b), and this framework enables a robust and smooth sensory reconstruction which facilitates learning and control.

\subsection{RL implementation}\label{section_RL_implementation}

As illustrated in Figure~\ref{Fig:Framework_Overview}a, our control objective is for the snake to reach a target location across a complex environment.
This process can be described by the following optimization problem
\begin{equation}\label{eq:optimization}
    \begin{aligned}
        & \max_{\mathbf{a}_t}\ \left[\sum_{j=1}^{N_L} \ r_b+ \sum_{i=1}^{N_R}r_{i}\cdot\mathds{1}\{d<D_{i}\}\right] \\
        \text{subject to:}&\quad \text{Equations~\ref{eq:lin} and \ref{eq:ang}}
    \end{aligned}
\end{equation}
where $N_L$ is the number of learning steps in one epoch, $d$ is the distance between the snake's head and the target, $r_b \propto \Delta d=d_{j}-d_{j+1}$ is the base reward promoting consistent progress toward the target, and $r_{i}$ denotes a set of additional incentives given when the snake enters one of $N_R$ predefined proximity zones (each of radius $D_{i}$) around the target.
Full details of the reward formulation are provided in Section~\ref{section_method_learningSetup}.
A RL framework is then leveraged to solve this optimization problem, integrating the actuation and sensory models defined above.

Described in Section~\ref{section_muscular_activation}, the snake's actions can be compactly defined by the two-wave motion template as $\mathbf{a}_t = \{A, \Phi\}$, simplifying the actuation of its distributed, continuum body.
Meanwhile, the state vector is defined as $\mathbf{s}_t = \{d, \alpha, \mathbf{v}, H_1, H_2\}$, where $\alpha$ represents the angle between the snake's bearing and the target, and $\mathbf{v}$ denotes the center-of-mass (COM) velocity.
Environmental sensing is encoded through $H_{1,2}$, derived from the oscillator outputs $\hat{h}_{1,2}$.
Here, $\hat{h}_{1,2}$ are not directly employed in $\mathbf{s}_t$ as they continuously oscillate over each undulation cycle, and their values at any single time point do not reflect the overall characteristics of the environment.
Therefore, we define and utilize $H_{1,2} = \frac{1}{T}\int_{t}^{t+T}|\hat{h}_{1,2}|\text{d}t$, the cycle-averaged force magnitudes, as the environmental sensing information.
All state and action variables are normalized to the range $[-1, 1]$ in accordance with standard RL practices.

%Overall, the integration of bio-inspired actuation and sensing models into the RL framework provides an effective abstraction for controlling a high-dimensional, continuum body. By reducing the system’s complexity to compact action and state spaces, this approach not only simplifies policy representation but also lowers computational demands during training and streamlines hyperparameter tuning.
\section{Navigation on flat surfaces} \label{Navigation_flat}

With the control framework defined, we outline the snake's learning procedure and navigation capabilities, starting with a simplified environment consisting of flat, uniform terrains.
To meaningfully represent the interfacial effects of snake locomotion observed in nature, we consider two representative frictional regimes: anisotropic ($\mu_t / \mu_f = 10$) and isotropic ($\mu_t / \mu_f = 1$) frictions~\cite{Hu:2009, Rieser:2021}.
Anisotropic friction arises from the snake's skin texture and uneven terrain features such as grass or rocks.
In contrast, isotropic friction captures the effect of deformable substrates---such as sand or mud---where the snake's inherent anisotropy is effectively neutralized due to substrate remodeling.
Snakes naturally inhabit both types of environments and are known to adopt distinct gait strategies to navigate effectively.
Therefore, to quantitatively understand snake behaviors in both cases and to streamline the subsequent learning process, we first conduct a pre-training step by testing a sweep of action parameters.

\subsection{Pre-training characterization}

\begin{figure}[!tb]
\centering
\includegraphics[width=0.99\textwidth]{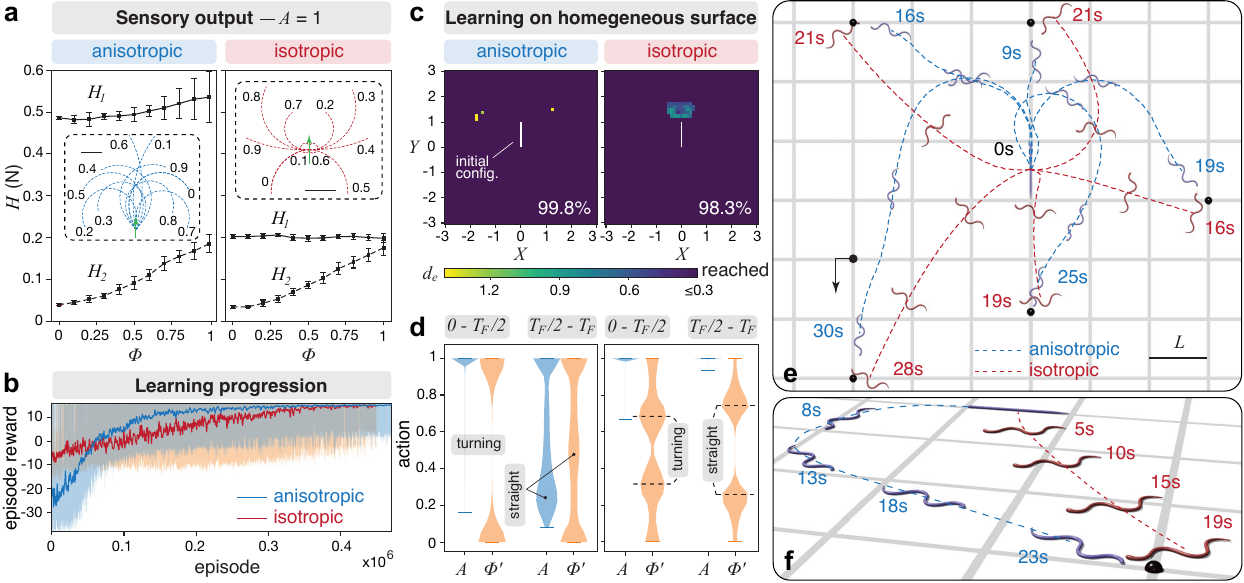}
\caption{
    \textbf{Sensing and learning on homogeneous surface.}
    \textbf{(a)} Sensory outputs when running open-loop simulations of snake on anisotropic and isotropic surfaces.
    Lifting amplitude $A=1$ is fixed for all runs and phase $\Phi$ varying within $[0,1]$.
    Simulation time $T_F=\qty{30}{\second}$, and each data point represents the mean and standard deviation of $H$ values for the last \qty{20}{\second} (when snakes stabilized).
    Insets demonstrate the snake's COM trajectory corresponding to each $\Phi$.
    \textbf{(b)} Progressions for learning on anisotropic and isotropic surfaces.
    The same set of hyperparameters is employed for both cases, with 10M total learning steps and a batch size of 16384.
    \textbf{(c)} Reachability heatmaps for learned polices showing snake-to-target distance at final time step $d_e$.
    2401 simulations are tested for each case, with each simulation assigned a target in a $49 \times 49$ grid.
    The maximum time allowed in each simulation is \qty{40}{\second}.
    Simulation terminates when the target is reached ($d_e<$0.3).
    Dimensions are normalized by the snake length $L$.
    \textbf{(d)} Distributions of the actions employed in all 2401 simulations in (c).
    Here, $T_F$ indicates the actual final simulation time for each simulation, and is $\leq \qty{40}{\second}$.
    \textbf{(e)} Top view comparison of anisotropic and isotropic snake gaits and trajectories for reaching 5 different targets using learned policies.
    The origin coincides with the snake's tail at $t=0$.
    \textbf{(f)} Perspective view comparison for trajectories reaching a target at (-3$L$,-$L$).
}\label{Fig:Result_Homo}
\end{figure}

The pre-training step is performed by simulating the snake with a set of constant action pairs, fixing $A = 1$ (maximum lift) and sweeping the phase offset $\Phi \in [0,1]$.
The resulting steady-state behaviors are shown in Figure~\ref{Fig:Result_Homo}a, which includes COM trajectories and the corresponding sensory feedback.
Under anisotropic friction, the snake exhibits a variety of forward slithering gaits, with its turning controlled by $\Phi$.
In contrast, when applying the same action sweep in the isotropic regime, the snake maintains maneuverability but transitions to a sidewinding gait---characterized by lateral translation relative to the body orientation.
This friction-dependent gait transition has been both experimentally observed and theoretically explained~\cite{Zhang:2021}.

Importantly, the snake's turning behavior does not vary uniformly with respect to $\Phi$ in both frictional regimes.
For instance, the extrema of $\Phi$ do not produce the sharpest turns, and within certain regions (e.g. $\Phi \in [0, 0.2]$), a small change in phase can lead to large deviation in the snake's trajectory.
These nonlinearities introduce additional challenges in learning a stable control policy.
To address this, we apply a linear remapping of the phase variable ($\Phi \mapsto \Phi'$) to regularize the relationship between control inputs and turning responses.
Details of the remapping are provided in the SI.
For consistency, the phase variable hereafter refers to $\Phi'$.

In addition to snake trajectories, Figure~\ref{Fig:Result_Homo}a also displays the average sensory outputs $H_{1,2}$ recorded during snake motion.
Reflecting the lateral forces experienced by the snake, the $H_1$ values demonstrate a clear and consistent distinction when the snake is deployed in different environments, despite different phases.
In contrast, vertical force measurements $H_2$ do not strongly depend on frictional isotropy/anisotropy; instead, they are primarily influenced by $\Phi'$.
These results confirm the complementary roles of $H_1$ and $H_2$ as effective sensory modalities that enable the snake's proprioception and environmental perception.

\subsection{Learning on homogeneous surfaces}\label{Section_Homo}
With insights from this initial characterization, we employ the Proximal Policy Optimization (PPO) algorithm~\cite{Schulman:2017} to learn the control policy.
In each training episode, the snake is initialized at the origin and tasked to reach a randomly placed target within a $6L \times 6L$ square domain.
To evaluate and compare the snake's learning performance across different environments, the same training procedure is conducted on both anisotropic and isotropic surfaces.
Details of the training parameters can be found in~\Cref{section_method_learningSetup}.

Figure~\ref{Fig:Result_Homo}b showcases the learning progress, demonstrating the convergence of both policies after $\sim 0.4$ million episodes.
The effectiveness of the learned policies is assessed through reachability tests, where the snake is assigned different targets.
Each simulation is terminated upon either successfully reaching the target or exceeding a maximum time limit.
At the end of each trial, we record the final snake-to-target distance $d_e$, and visualize it as spatial heatmaps in Figure~\ref{Fig:Result_Homo}c.
As shown in the plots, the learned policies enable the snake to reach nearly all points within the domain, achieving close to 100\% success rates in both frictional environments.
We note that in the isotropic case, a small hard-to-reach region persists right in front of the snake.
This is primarily due to the snake's sideways motion (sidewinding), which costs the snake extra time to reorient.
%This limitation could be addressed by incorporating body extensions, such as the striking motions.

To further evaluate the learned policies, we examine the distribution of action archetypes during the reachability tests by binning actions according to when they occur within each trial: in the first versus second half of the simulation horizon $T_F$ (Figure~\ref{Fig:Result_Homo}d). This temporal segmentation reveals motion sequences of apparent intention. In the anisotropic setting, snakes first focus on reorienting their body toward the target, accomplished by activating lift and locking the phase near its extrema to maximize turning (see the remapped phase space in the SI).
Then, they markedly reduce lift and shift the phase to favor straight trajectories, thereby closing the remaining distance. A similar two-stage pattern emerges in the isotropic setting, with key differences that lift remains persistently engaged to sustain propulsion in this environment~\cite{Zhang:2021}, and a smaller phase shift is sufficient to transition from turning to straight sidewinding motions (as informed by Figure~\ref{Fig:Result_Homo}a). These two-stage patterns are further validated by visualizing the snake's COM trajectories (Figure~\ref{Fig:Result_Homo}e).

%To further assess the performance of the learned policies, we analyze the distribution of action archetypes during the reachability tests, by categorizing them based on their time of occurrence within each trial---specifically, whether they fall within the first or second half of the simulation duration $T_F$ (Figure~\ref{Fig:Result_Homo}d). This temporal segmentation reveals a motion sequence with apparent intentions, particularly for reaching distant targets.  In the anisotropic case, snakes predominantly focus on aligning their direction toward the target during the first half of the simulation, achieved by activating lift and locking the phase near its extrema to maximize turning (see the remapped phase space in the SI). In the second half, most snakes substantially reduce lift and adjust the phase to promote straight-line motion and close the remaining distance. Similar behaviors are also observed in the isotropic case, with the distinction that lift remains consistently activated to facilitate motion in this environment~\cite{Zhang:2021}.

In both environments, the learned policies enable precise turning maneuvers, allowing the snake to reach targets distributed across the domain.
Moreover, the results align with the expectation that forward slithering gaits facilitate faster progression toward frontal targets, whereas sidewinding is more advantageous for reaching targets positioned laterally or behind the snake (Figure~\ref{Fig:Result_Homo}f).
Notably, the snake's trajectories exhibit minimal wasteful detours, although traveled distance is not specifically penalized in Equation~\ref{eq:optimization}.
We attribute this outcome to the compact yet expressive action space, which promotes effective locomotion.

\subsection{Policy switch on heterogeneous surfaces}\label{Section_Heter}

Having established that the snake can successfully learn to navigate homogeneous environments, we extend our study to a heterogeneous setting with regions of different frictional properties. Taking advantage of previously learned policies, we tackle this scenario by considering a policy-switching approach, where the snake dynamically selects between learned policies based on its sensed environment.

%With the snake successfully learning how to navigate in homogeneous environments, we now extend our study to a more complex, heterogeneous scenario composed of different frictional regions. Taking advantage of previously learned policies, we tackle this problem by considering a policy-switching approach, where the snake dynamically selects between learned policies based on its environmental sensing capability.

\begin{figure}[!tb]
\centering
\includegraphics[width=0.99\textwidth]{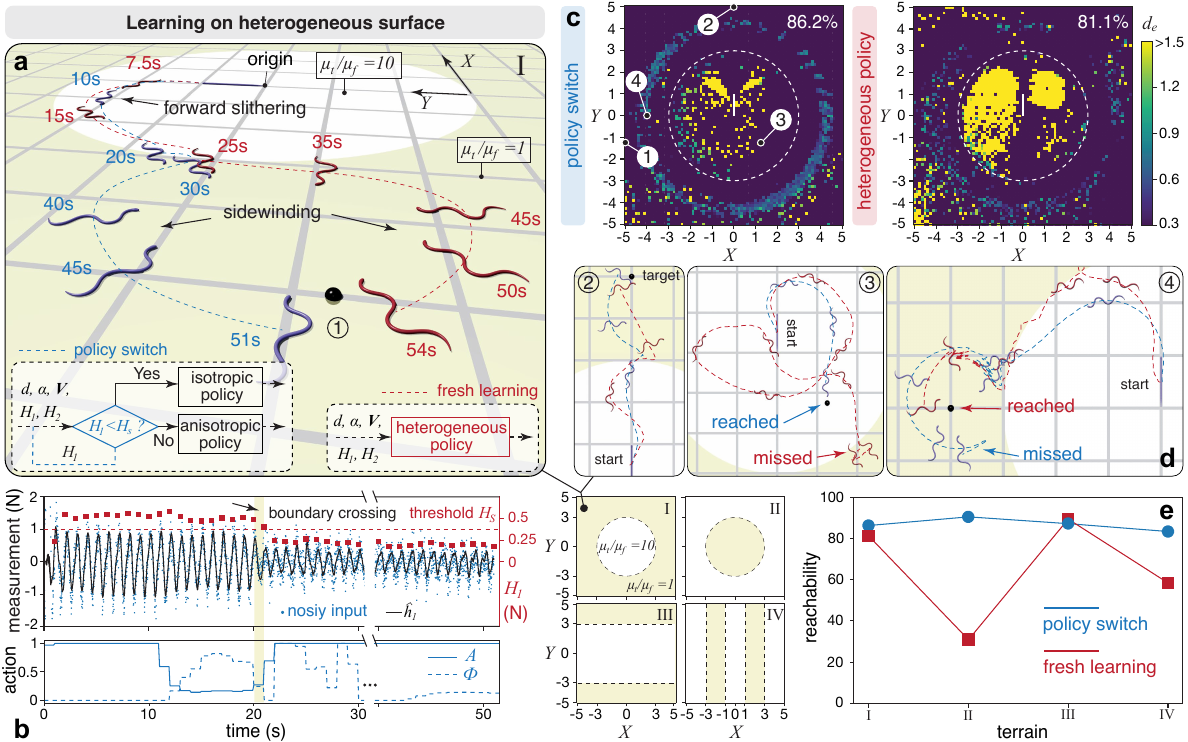}
\caption{
    \textbf{Navigation on heterogeneous frictional surfaces through policy switch.}
    \textbf{(a)} Comparing snakes' trajectories and gait transitions on a heterogeneous surface (court I.) using two different control strategies: policy switch between two homogeneous polices and freshly learned heterogeneous policy.
    The surface contains a circular anisotropic frictional pattern around the center, with radius of 3$L$.
    The rest of the surface is instead characterized by isotropic friction.
    Target is at (-5$L$,-1.25$L$).
    \textbf{(b)} Sensory outputs and actions of the snake for reaching the target in (a) under the guidance of the policy-switching approach.
    Blue dots indicate the noisy ground force measurements (input of FPF), while black line and red squares are the filtered signal and state feedback, respectively.
    The threshold value $H_{s}$ is set to be \qty{0.375}{\newton}.
    \qty{10}{\second} of the plot during sidewinding are trimmed for the interest of space.
    \textbf{(c)} Reachability tests for two control strategies.
    6561 simulations are tested for each case, with each simulation assigned a target in a $81 \times 81$ grid.
    Termination time for each simulation is at \qty{80}{\second}.
    Dimensions are normalized by $L$.
    \textbf{(d)} Three representative test cases showcasing snake behaviors in reaching targets at different locations.
    Targets are marked at corresponding locations in the heatmap in (c).
    \textbf{(e)} Testing the robustness of policy switch in 3 additional courts: court II (reverse of court I), court III (anisotropic strip along $X$ axis) and court IV (double isotropic strips along $Y$ axis).
    Plot showcasing the comparison of the reachability between policy switch and fresh learning for each court.
    Shaded region indicates the upper and lower bound reachabilities by deploying policy mix-match: running freshly learned policy from one court on other 3 courts.
    Details about the policy mix-match results can be found in the SI.
}\label{Fig:Result_Heter}
\end{figure}

To evaluate this strategy, we construct a heterogeneous terrain consisting of a circular anisotropic frictional region embedded within an otherwise isotropic domain (Figure~\ref{Fig:Result_Heter}a). The snake is initialized at the center of the anisotropic zone and must transition between different gaits to reach a target located outside the circle. This behavior is enabled by a simple decision-making mechanism based on the value of $H_1$, which serves as a key indicator of the local frictional environment, as characterized in Figure~\ref{Fig:Result_Homo}a. Our observations show that $H_1$ remains relatively constant within a given frictional surface and changes only when the snake transitions between distinct surfaces. Consequently, a threshold value $H_s$ can be used to distinguish between frictional environments and guide policy selection. At each step, the current value of $H_1$ is compared against the threshold $H_s$ (Figure~\ref{Fig:Result_Heter}b), defined as the mean $H_1$ across the two homogeneous environments. If $H_1 < H_s$, the terrain is identified as isotropic and the corresponding locomotion policy is activated; otherwise, the anisotropic policy is selected. The resulting trajectory, shown in Figure~\ref{Fig:Result_Heter}a, illustrates the successful deployment of this strategy: the snake accurately transitions between gaits while traversing the heterogeneous terrain and ultimately reaches the target. The associated sensory measurements and policy selections are presented in Figure~\ref{Fig:Result_Heter}b.

%To test this strategy, we design a heterogeneous terrain featuring a circular anisotropic frictional zone embedded within an otherwise isotropic domain (Figure~\ref{Fig:Result_Heter}a). The snake is initialized at the center of the anisotropic region and is required to transition between different gaits to reach a target outside the circle. This is enabled through a simple decision-making strategy based on the value of $H_1$, the key indicator of environmental friction, as characterized in Figure~\ref{Fig:Result_Homo}a. Based on our observations of $H_1$, it remains relatively constant in the same frictional surface and only changes value when it enters a different frictional surface. Hence, a simple threshold $H_s$ can distinguish between frictional environments and be used to select between policies. At each step, the current $H_1$ is compared to a threshold $H_s$ (Figure~\ref{Fig:Result_Heter}b), which was chosen to be the mean $H_1$ value across the two homogeneous environments. If $H_1<H_s$, then the snake detects an isotropic terrain and activates the corresponding policy; otherwise, it will default to activating the anisotropic policy. The snake trajectory in Figure~\ref{Fig:Result_Heter}a showcases the deployment of this approach, demonstrating a successful gait transition as the snake navigates the heterogeneous court and reaches the target. The corresponding sensory outputs and action selections are plotted in Figure~\ref{Fig:Result_Heter}b.

To further characterize the effectiveness of this approach, we compare it with a policy trained from scratch on the same heterogeneous terrain using an identical training protocol as in the homogeneous setting. The reachability results in Figure~\ref{Fig:Result_Heter}c indicate comparable performance between the two methods, with the policy-switching approach achieving a slightly higher success rate (86.2\% vs. 81.1\%). In many cases, both strategies exhibit similar gait transitions, enabling the snake to successfully reach the target, as illustrated in Figure~\ref{Fig:Result_Heter}a and Figure~\ref{Fig:Result_Heter}d \textcircled{2}. However, each method also demonstrates distinct limitations. The policy-switching approach struggles with targets located just outside the anisotropic region, where smooth transitions between gaits are desirable but not explicitly learned (see Figure~\ref{Fig:Result_Heter}d \textcircled{4}). In contrast, the policy trained from scratch shows notably weaker performance within the anisotropic region (Figure~\ref{Fig:Result_Heter}d \textcircled{3}), suggesting that the increased environmental complexity poses greater challenges during training.

%To further evaluate the effectiveness of this approach, we compare it against a policy trained from scratch on this heterogeneous surface with the same training protocol as in the homogeneous setting. The reachability tests shown in Figure~\ref{Fig:Result_Heter}c reveal comparable reachability between the two approaches, with policy-switching outperforming fresh learning by a small margin (86.2\% vs 81.1\%). In most cases, both strategies guide similar gait transitions, allowing the snake to successfully reach targets, such as in Figure~\ref{Fig:Result_Heter}a and Figure~\ref{Fig:Result_Heter}d \textcircled{2}. However, the two strategies also exhibit distinct limitations: the policy-switching approach struggles with targets just outside the anisotropic region, where smooth gait transitions are required but not explicitly trained for (exemplified in Figure~\ref{Fig:Result_Heter}d \textcircled{4}). Meanwhile, the freshly learned policy substantially underperforms within the anisotropic domain (Figure~\ref{Fig:Result_Heter}d \textcircled{3}), suggesting an overall increased training complexity arising from the more challenging terrains.

However, the policy-switching approach demonstrates superior overall performance when evaluated across multiple heterogeneous terrains with varying features, as shown in Figure~\ref{Fig:Result_Heter}e. In all cases, its reachability remains indeed consistently high, either matching (cases I and III) or substantially exceeding (cases II and IV) the performance of policies trained from scratch. We note that the freshly learned policies follow the same training protocol as in the homogeneous setting and could potentially benefit from further hyperparameter tuning or extended training duration. Nevertheless, these results highlight the policy-switching approach as an efficient and scalable solution for adaptive snake locomotion across diverse heterogeneous environments, effectively eliminating the need for retraining in each new setting.

%The robustness of the policy-switching approach is further assessed by deploying it across three additional heterogeneous terrains (Figure~\ref{Fig:Result_Heter}e). In all cases, its reachability remains consistently high, closely matching or even surpassing the performance of freshly trained policies. In contrast, freshly learned policies suffer from limited generalizability, as substantial performance degradation is observed in the policy mix-matching tests, where a policy trained on one court is deployed on different terrains (shaded regions in Figure~\ref{Fig:Result_Heter}e; details provided in SI Figure). We note here again that the fresh learning procedures follow the same setup as in the homogeneous case and could potentially be improved through further hyperparameter tuning or extended training. Nevertheless, these results emphasize the policy-switching approach as a scalable and energy-efficient solution for adaptive snake control across diverse heterogeneous environments, eliminating the need for retraining in each new setting.
\section{Navigation on 3D heterogeneous terrains}\label{Navigation_Mesh}

Building on these findings, and motivated by the need to more faithfully capture the complexity of natural snake environments, we extend our investigation to encompass realistic three-dimensional terrains.

\subsection{Policy adaptation}\label{Section_head_raising}

\begin{figure}[!tb]
\centering
\includegraphics[width=0.99\textwidth]{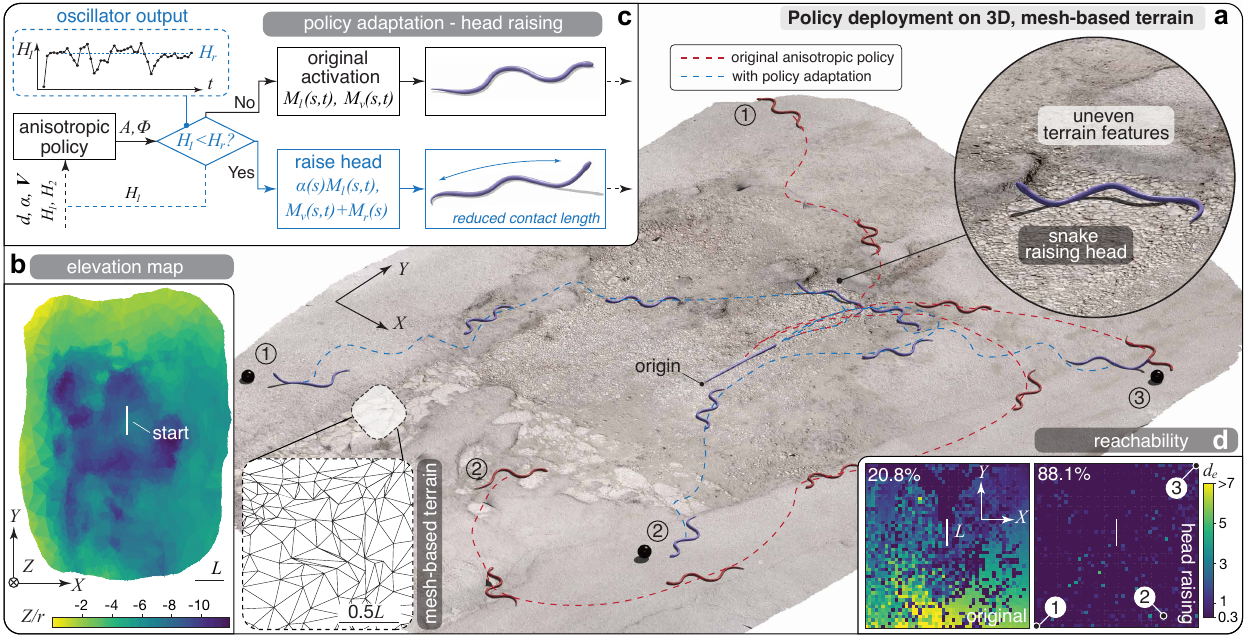}
\caption{
    \textbf{Deploying learned policy on 3D, mesh-based terrain.}
    \textbf{(a)} Snake moving on an asphalt ground, characterized by anisotropic friction ($\mu_t / \mu_f = 10$).
    Trajectories showcase snake navigation with (blue) and without (red) policy adaptation.
    The snake is initialized with its tail aligned with the origin of the court.
    Three targets are at (-3$L$,-3$L$), (1.75$L$,-2.5$L$) and (3$L$,3$L$), respectively.
    Inset at the top right corner of the figure showcases the instantaneous head raising of the snake when encountering uneven terrain features such as a bump.
    Inset at the bottom of the figure demonstrates the triangular meshes that constitute a small piece of the surface.
    \textbf{(b)} Elevation map showcasing the 3D terrain features of the court.
    Dimension in the $Z$ direction is normalized by snake radius $r$.
    \textbf{(c)} Diagram of policy adaptation based on sensing of the lateral impulses $H_1$.
    The inset plot demonstrates an example of the sensory output with values momentarily below the head raising threshold $H_r$, which is determined as \qty{0.4375}{\newton} based on the pre-training data.
    \textbf{(d)} Comparison of the snake's reachability with and without policy adaptation.
    2401 simulations are tested for each case, with each simulation assigning a target in a $49 \times 49$ grid.
    The simulation termination time is set to \qty{60}{\second}.
}\label{Fig:Mesh_Simple}
\end{figure}

%To this end, we continue leveraging our previously trained policy to assess whether learning in simplified environments, together with the passive adaptability afforded by the snake's compliant body, can enable effective traversal in complex settings without additional training. To test this idea, we deploy the snake on an anisotropic real-world terrain (Figure~\ref{Fig:Mesh_Simple}a; model details provided in the SI) that is predominantly flat but contains irregular features such as bumps and cracks, as evidenced by the elevation map in Figure~\ref{Fig:Mesh_Simple}b.  After deploying the trained anisotropic policy and performing the same reachability test described in \Cref{Section_Homo}, we observe a poor success rate of 20.8\%, with the snake missing the majority of targets behind its initial position. Although the snake initiates appropriate turning motions at the onset, terrain irregularities intermittently disrupt ground contact and reduce turning momentum, ultimately compromising maneuverability (see, e.g., the red trajectories attempting targets \textcircled{1} and \textcircled{2}).

To this end, we continue leveraging our previously trained policy to assess whether learning in simplified environments, combined with the passive adaptability afforded by the snake's compliant body, can enable effective traversal in more complex settings without additional training. To evaluate this hypothesis, we deploy the snake on a realistic anisotropic terrain (Figure~\ref{Fig:Mesh_Simple}a; model details provided in the SI) that is predominantly flat but features irregularities such as bumps and cracks, as illustrated by the elevation map of Figure~\ref{Fig:Mesh_Simple}b.

Direct application of the previously trained anisotropic policy results in a low success rate (20.8\% in the reachability test described in \Cref{Section_Homo}), with the snake failing to reach most targets located behind its initial pose. Although the snake initiates appropriate turning motions at the outset, terrain irregularities intermittently disrupt ground contact and degrade maneuverability.
Despite these limitations, the simulations reveal promising indications that motivate further refinement of this approach. Inspired by biological snakes---which naturally exploit transient three-dimensional behaviors to negotiate complex terrains---we introduce a head-raising mechanism that enables the snake to temporarily lift the anterior portion of its body. This modification effectively shortens the ground contact length, increasing traction forces, thereby enhancing turning agility.
The activation of this head-raising mechanism is once again governed by the sensory output $H_1$. In this case, we adopt a newly defined threshold $H_r$ ($H_r > H_s$), set to the lower bound of typical $H_1$ values observed on anisotropic surfaces (as inferred from Figure~\ref{Fig:Result_Homo}a). By default, the snake follows the lateral ($M_l$) and vertical ($M_v$) activations prescribed by the anisotropic policy. When $H_1 < H_r$, indicating a transient loss of effective ground contact due to surface asperities, an additional muscular input $M_r$ is activated to raise the head, while the lateral activation of the elevated segment is attenuated via a modulation function $\alpha(s)$, with $\alpha(s) < 1$, to avoid unnecessary oscillation. A schematic of this adaptation is shown in Figure~\ref{Fig:Mesh_Simple}c, with full implementation details provided in the SI.

After incorporating the head-raising mechanism, we repeat the same target-reaching task to evaluate performance. As illustrated by the blue trajectories in Figure~\ref{Fig:Mesh_Simple}a, head-raising allows the snake to execute rapid trajectory adjustments, effectively realigning its motion and consequently reaching all assigned targets. The reachability test shows a substantial increase in success rate---from 20.8\% to 88.1\% (Figure~\ref{Fig:Mesh_Simple}d). These results highlight the effectiveness of this bio-inspired policy adaptation, which integrates active sensing with transient three-dimensional behaviors to provide a computationally lightweight, yet robust solution for traversing realistic 3D terrains.

\subsection{Deployment in complex environments}

\begin{figure}[!tb]
\centering
\includegraphics[width=0.99\textwidth]{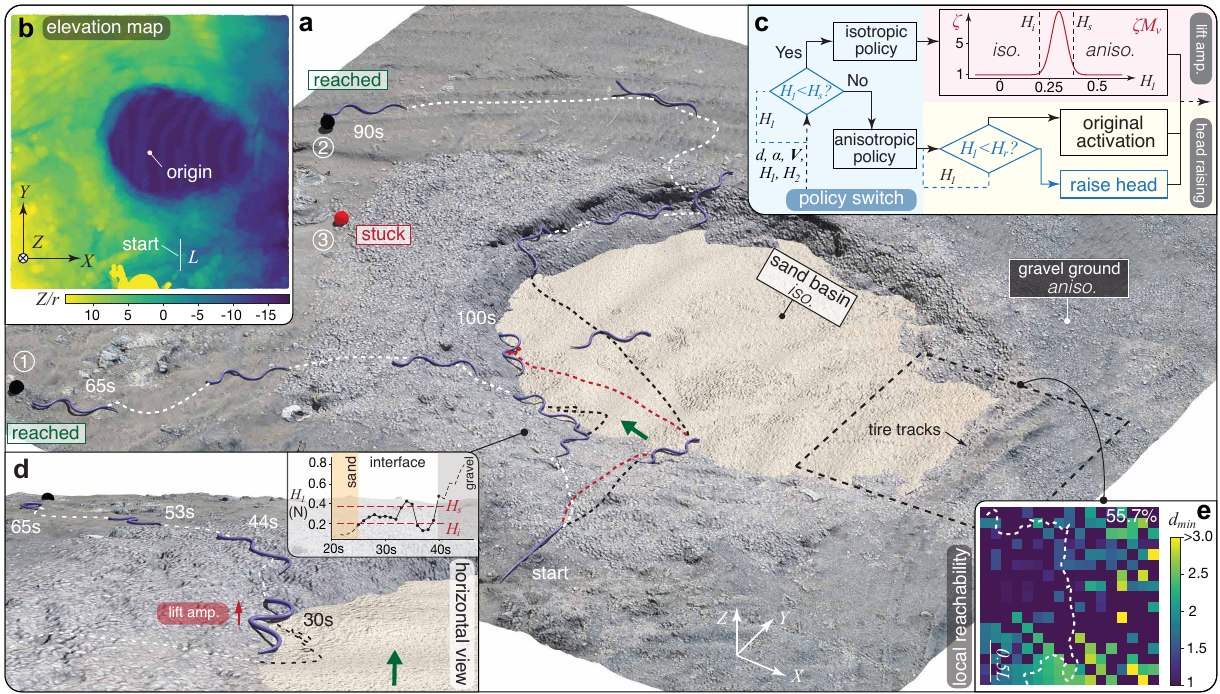}
\caption{
    \textbf{Snake navigation in complex 3D environments.}
    \textbf{(a)} A complex environment comprised of two types of terrains: a gravel ground with anisotropic friction, and a sandy surface characterized by isotropic friction. Three examples demonstrating the snake's capability of navigating the environment while switching between two locomotion strategies.
    The snake successfully reaches the target located at (-3.9375$L$, 1.1875$L$) and (-3.75$L$, -3.6875$L$), but fails to reach the target at (-3$L$, 0).
    The snake is initialized with its tail at (1$L$,-3.75$L$).
    \textbf{(b)} Elevation map showcasing the complex terrain features and the height difference between the two types of grounds.
    The $Z$ coordinate is normalized by snake radius $r$.
    \textbf{(c)} The overall control framework integrating policy switch and two different policy adaptation strategies, lift amplification and head raising, enabling snakes traversing complex 3D terrains.
    Plot in the lift amplification block demonstrates the variation of the amplification factor $\zeta$ given different sensory output of $H_1$.
    $\zeta$ is defined as a scaled normal distribution: $\zeta = 1+3/(\sigma\sqrt{2\pi}) \exp\left[-2{(H_1-0.3)}^2/\sigma^2\right]$, where $\sigma=0.15$ restricts the span of amplification in between $H_i$ and $H_s$.
    \textbf{(d)} Zoomed-in view of the snake leveraging lift amplification to overcome the height difference while making transition from the sand basin to the gravel ground.
    The inset plot showcasing the $H_1$ values during the transition period.
    The green arrow denotes the viewing direction.
    \textbf{(e)} Characterization of the snake's reachability to a square region on the right side of environment, bounded by four corners at (2.125$L$,-1$L$), (2.125$L$,1$L$), (4.125$L$,-1$L$) and (4.125$L$,1$L$), respectively.
    Heatmap plots the minimum snake-to-target distance $d_{\min}$ in each simulation.
    The white dashed line indicates the boundary of the two terrains.
    289 simulations are tested with each simulation assigned a target in a $17 \times 17$ grid.
    Termination time for each simulation is at \qty{100}{\second}.
}\label{Fig:Mesh_Switch}
\end{figure}

Finally, we challenge our control strategy on a highly heterogeneous, uneven environment.

As illustrated in Figure~\ref{Fig:Mesh_Switch}a, this environment combines both anisotropic and isotropic substrates, with gravel covering most of the domain and a sandy basin positioned at the center.
Beyond diverse substrate properties, this landscape also introduces pronounced topographical features, with elevation changes exceeding 30 times of snake's radius ($r$) and steep cliffs at the basin's boundary rising over $15r$ (Figure~\ref{Fig:Mesh_Switch}b).
Therefore, successfully traversing this landscape requires the snake to not only switch between different gaits and negotiate ground asperities, but also to overcome substantial height difference.
Having addressed the first two challenges in our previous investigations, we now introduce a final policy adaptation to facilitate obstacle climbing.

With the most significant obstacles being the cliff between the two terrain types, transitioning from the sandy surface to the gravel ground presents a key challenge.
To address this, we apply a new form of policy adaptation to the isotropic policy.
Specifically, we introduce a new threshold $H_i$, representing the upper bound of typical $H_1$ values measured on the isotropic terrain (Figure~\ref{Fig:Result_Homo}a).
This adaptation is activated during the snake's transition phase---when $H_i < H_1 < H_s$---indicating that the snake, while still on the sandy surface, is beginning to encounter the gravel terrain and its associated obstacles.
In response, the snake gradually increases the amplitude of the lift wave along its entire body by a factor $\zeta$, allowing it to raise itself over the obstacles.
Meanwhile, $\Phi$ continues to follow the original value issued by the isotropic policy to maintain correct directionality. %Detailed parameters are provided in the Experimental Section.

The implementation of this policy adaptation is illustrated in Figure~\ref{Fig:Mesh_Switch}c, which also outlines the complete decision-making architecture.
It begins with the policy-switching mechanism introduced in Section~\ref{Section_Heter}, which selects between the isotropic and anisotropic policies based on the environmental sensing result.
Each policy then incorporates its respective adaptation: lift amplification within the isotropic policy and head raising within the anisotropic policy.
The effectiveness of this integrated control strategy is demonstrated in Figure~\ref{Fig:Mesh_Switch}a through two representative navigation trials.
In reaching target \textcircled{1}, the snake initially detects the gravel terrain and employs the anisotropic policy to perform a turning maneuver toward the target.
As the snake briefly enters the sandy region along the terrain boundary, the lift amplification is triggered, enabling it to successfully overcome the adjacent cliff (Figure~\ref{Fig:Mesh_Switch}d) before returning to gravel and ultimately reaching the target.
The trajectory toward target \textcircled{2} further highlights robust gait transitions in this complex scenario, as the snake stably executes the sidewinding motion across the sandy area.
During the final approach, snapshots also capture instances of head raising that facilitate sharp realignments and successful target acquisition.

Although these examples demonstrate successful deployments of our integrated control strategy, the highly heterogeneous and topographically complex nature of this environment makes reliable navigation inherently difficult, and thus failures are expected.
For instance, the attempt to reach target \textcircled{3} is unsuccessful due to a relatively smoother cliff at the interface between the two terrain types, where the snake loses traction despite lift amplification.
Additionally, a reachability test conducted over a section of the terrain reveals that while most of the snakes (55.7\%) could approach the target within 1$L$ (Figure~\ref{Fig:Mesh_Switch}e), the reachability remains low at $\sim30\%$.
This performance drop is primarily attributed to the difficulty of reaching targets located along the interface between the two terrains, a challenge already inherent to the heterogeneous flat-ground setting shown in Figure~\ref{Fig:Result_Heter}c. Here, complex terrain asperities further exacerbate this challenge, making target reachability during gait transitions particularly difficult.

These findings underscore the limitations of the current framework, which relies on learning in simplified environments and policy adaptations that do not substantially modify the underlying actuation patterns. Nevertheless, our strategy establishes a computationally efficient baseline for generalizing snake locomotion across diverse environments. To overcome highly variable 3D terrains, future efforts may focus on dedicated training for specific features, or the incorporation of task-specific motor primitives—such as biologically inspired gaits for climbing, reaching, or anchoring—that better capture the full behavioral repertoire observed in natural snakes. The proposed framework provides a modular foundation that can readily accommodate such additional strategies.
\section{Conclusion}

In this work, we examine how soft, limbless bodies can learn to navigate complex landscapes using a Cosserat rod-based snake model operating in realistic 3D environments. We propose a reinforcement learning-based control strategy that integrates bio-inspired actuation and sensing to manage the high degrees of freedom inherent in continuum systems.
Initially validated in simplified, homogeneous settings, the framework produces locomotion policies that achieve precise and robust navigation. These learned behaviors are then extended to more challenging scenarios through a lightweight decision-making mechanism that coordinates multiple policies, enabling adaptive snake-like behaviors without additional training.
This modular architecture promotes generalization across heterogeneous substrates and supports deployment in topographically complex terrains, where modest bio-inspired policy adjustments further enhance performance. Overall, our approach provides a scalable and computationally efficient method for controlling limbless locomotion, offering both a high-fidelity platform for studying snake behavior in natural environments and practical insights for the design and deployment of soft, slithering robots in real-world applications.

%In this work, we investigate how soft limbless bodies can learn to navigate complex landscapes by deploying a Cosserat rod-based snake model in realistic 3D environments. A control strategy based on reinforcement learning is proposed, integrating bio-inspired actuation and sensing mechanisms to effectively manage the high-DOF inherent to soft, continuum bodies. First demonstrated in simplified, homogeneous settings, this learning framework yields effective locomotion policies capable of precise and robust navigation. These policies are then extended to more complex scenarios, coordinated through a lightweight decision-making mechanism that enables adaptive snake behaviors without additional training. This modular architecture facilitates generalization across heterogeneous substrates and supports deployment in topographically complex terrains, where modest, bio-inspired policy adaptations further enhance performance. Taken together, our investigation offers a scalable and computationally efficient approach to controlling limbless locomotion, providing both a high-fidelity platform for studying complex snake behavior in natural environments and practical guidance for deploying soft slithering robots in real-world applications.

\newpage
\section{Experimental section}\label{Material_Methods}

\subsection{Cosserat rod model of the snake}\label{section_method_CosseratRod}

As shown in Figure~\ref{Fig:Framework_Overview}b, the snake is modeled as a Cosserat rod to capture large body deformations arising from internal actuation and environmental interaction. The rod is described by a centerline $\bar{\mathbf{x}}(s, t) \in \mathbb{R}^3$ and an orthonormal material frame $\mathbf{Q}(s, t) = \{\bar{\mathbf{d}}_1, \bar{\mathbf{d}}_2, \bar{\mathbf{d}}_3\}^{-1}$, where each $\bar{\mathbf{d}}_i$ represents a director in the local frame at point $s$. Any vector $\bar{\mathbf{v}}$ in the global (lab) frame can then be transformed into its corresponding local representation $\mathbf{v}$ via $\mathbf{v} = \mathbf{Q} \bar{\mathbf{v}}$, with the inverse transformation $\bar{\mathbf{v}} = \mathbf{Q}^T \mathbf{v}$. Here, $s \in [0, L_c(t)]$ is the arc-length coordinate, where $L_c(t)$ denotes the rod's current length. In the absence of shear, $\bar{\mathbf{d}}_3$ aligns with the tangent vector $\bar{\mathbf{x}}_s = \partial_s \bar{\mathbf{x}}$, while $\bar{\mathbf{d}}_1$ and $\bar{\mathbf{d}}_2$ span the normal-binormal plane. Shear and stretch decouple $\bar{\mathbf{d}}_3$ from the tangent vector, resulting in the shear vector $\boldsymbol{\sigma} = \mathbf{Q} (\bar{\mathbf{x}}_s - \bar{\mathbf{d}}_3)$. The curvature of the rod $\boldsymbol{\kappa}$ encodes the rotation rate of the material frame along the body and satisfies $\partial_s \mathbf{d}_j = \boldsymbol{\kappa} \times \mathbf{d}_j$, while the angular velocity is given by $\partial_t \mathbf{d}_j = \boldsymbol{\omega} \times \mathbf{d}_j$. The linear velocity of the centerline is $\bar{\mathbf{v}} = \partial_t \bar{\mathbf{x}}$. Furthermore, the geometry of rod can be characterized by the second moment of area $\mathbf{I}$ and cross-sectional area $A$, with density denoted by $\rho$. Together, these definitions yield the governing equations of a Cosserat rod, as described in~\cite{Gazzola:2018}:

\vspace{-10pt}
\begin{equation}
\partial_t^2 (\rho A \bar{\mathbf{x}}) = \partial_s (\mathbf{Q}^T \mathbf{n}) + \bar{\mathbf{F}}\label{eq:lin}
\end{equation}
\begin{equation}
\partial_t (\rho \mathbf{I} \boldsymbol{\omega}) = \partial_s \boldsymbol{\tau} + \boldsymbol{\kappa} \times \boldsymbol{\tau} + (\mathbf{Q} \bar{\mathbf{x}}_s \times \mathbf{n}) + (\rho \mathbf{I} \boldsymbol{\omega}) \times \boldsymbol{\omega} + \mathbf{Q} \bar{\mathbf{C}}\label{eq:ang}
\end{equation}
where Equations~\ref{eq:lin} and~\ref{eq:ang} represent the change of the linear and angular momentum at every cross-section, respectively. The terms $\mathbf{n}$ and $\boldsymbol{\tau}$ denote the internal forces and couples per unit length, while $\bar{\mathbf{F}}$ and $\bar{\mathbf{C}}$ are the external forces and couples per unit length, respectively.

To enable numerical simulation, Equations~\ref{eq:lin} and~\ref{eq:ang} are discretized into $(n+1)$ nodes connected by $n$ cylindrical elements. Linear displacements are governed by internal and external forces acting at the nodes, while angular rotations are captured through couples applied to the elements. In our snake simulations, the rod is subjected to muscular activation (Section~\ref{section_muscular_activation}), gravitational forces, ground contact and friction (SI), as well as viscous dissipation modeled via Rayleigh potentials. The resulting dynamics are computed through time integration of the discrete rod model using a second-order position Verlet scheme. Full implementation details can be found in~\cite{Gazzola:2018}.

To match the morphology of a corn snake, we instantiate the rod with a rest length of $L = 0.8$m, radius $R = 0.009$m, and total mass $m = 203.58$g. We note that during snake locomotion, bending is the predominant mode of deformation and axial stretch is negligible ($L_c(t)\approx L$). Therefore, $L$ is employed as the unit length for travel distance throughout this study. Moreover, the snake's dynamics are further governed by the ratio of inertia to friction forces, via the Froude number $Fr=(L/T^2)/(g \mu_f)$, where $g$ denotes gravitational acceleration. Throughout this study, we set $Fr = 0.1$ to capture the friction-dominated regime commonly observed in nature~\cite{Zhang:2021}. A full list of simulation parameters is provided in the SI.

\subsection{Coupled oscillator Feedback Particle Filter (FPF)}\label{section_method_oscillators}

The snake's environmental sensing is enabled by implementing the coupled oscillator FPF. Specifically, we employ 2 independent oscillator groups responsible for sensing the net lateral ($F_l$) and vertical ($F_v$) contact forces experienced by the central 20\% of the snake body. Each group contains $N=100$ oscillators, with each oscillator characterized by a phase variable $\theta$. The dynamics of the $i$-th oscillator in group $j$ are expressed as:
\begin{equation}
\text{d}\theta^i_{j}(t)=\omega^i_j\text{d}t+\frac{K_j(\theta_j^i(t),t)}{\sigma^2_w} \circ \left(\text{d}Z_j(t)- \frac{h_j(\theta_j^i(t),\mathbf{r}_j(t)) + \hat{h}_j(t)}{2}\text{d}t\right),
\label{oscillatordynamcis}
\end{equation}
where $\omega^i_{j}$ denotes the baseline angular velocity of the oscillator, $K_j$ is the gain function~\cite{Tilton:2013}, and $\circ$ indicates Stratonovich integration. The measurement process is defined as $\text{d}Z_j(t)=F_{l,v}\text{d}t+\sigma_w\text{d}W_j(t)$ ($F_l$ for $j=1$, $F_v$ for $j=2$), where $W_j(t)$ is a standard Wiener process describing the noise intrinsic to the measurement process and $\sigma_w$ is the
standard deviation. We note that the noise of $\text{d}Z$ is further compounded by the unsteady nature of $F_{l,v}$, due to the snake's oscillatory contact with uneven substrates. The observation function $h_j$ is numerically constructed through a linear combination of the Fourier basis function as $h_j(\theta_j^i) \approx \mathbf{r}^T \bm{\phi}_h(\theta_j^i)$, where $\bm{\phi}_h$ is a vector of $M_h$ dimensional Fourier basis, and $\mathbf{r} \in \mathbb{R}^{M_h}$ is a weight vector that is updated online according to~\cite{Wang:2020}. The group-level feedback is computed as the mean observation estimate across oscillators, defined as $\hat{h}_j(t)\approx N^{-1}\sum_{i=1}^Nh_j(\theta_j^i(t),\mathbf{r}_j(t))$. Additional implementation details and parameter specifications for the coupled oscillator FPF are provided in \cite{Tilton:2013, Wang:2020} and in the SI.

\subsection{RL implementation}\label{section_method_learningSetup}

We employ the PPO algorithm for RL training, implemented using the open-source Python package Stable Baselines 3~\cite{Raffin:2021}. Each policy is trained using 64 parallel environments on the Pittsburgh Supercomputing Center's Bridges-2 cluster. In each training environment, a single snake is initialized with its tail aligned at the center of the court, and a random target location is assigned. The simulation time and court dimensions vary by frictional environment and are specified in the captions of Figure~\ref{Fig:Result_Homo} and Figure~\ref{Fig:Result_Heter}. We note that for heterogeneous terrain cases, both the simulation time and court size are scaled to ensure sufficient experience gathered across both isotropic and anisotropic regions. After each episode, the snake is reset with a new, randomly selected target.

The training time step is set equal to the snake's undulation period $T = 1$s, allowing the snake to execute each action for a full cycle before receiving the next instruction. Overall, training is conducted for 10 million time steps. To optimize learning efficiency, multiple batch sizes in the range $[8192, 65536]$ are evaluated, with a batch size of 16384 ultimately selected based on overall performance.

The reward function is defined in Equation~\ref{eq:optimization}, where $r_b = 10 \Delta d$ is the base reward, the number of proximity zones $N_R$ is set to be 3, and the additional reward $r_i$ is increased as the snake gets closer: $(r_{i},D_{i})=\{(1,1.5L),(5,0.75L),(10,0.3L)\}$. We note that each additional reward is granted only upon the snake's first entry into a given proximity zone. A successful target reach is defined as achieving a final snake-to-target distance $d_e<0.3L$, determined according to a biologically plausible striking range.

\medskip
\noindent \textbf{Supporting Information.} Supporting Information is available from the authors.

\medskip
\noindent \textbf{Conflict of Interest.} The authors declare no conflict of interest.

%\medskip
%\noindent \textbf{Data Availability Statement.}

\medskip
\noindent \textbf{Acknowledgements.} This study is jointly funded by National Science Foundation (NSF) EFRI C3 SoRo \#1830881 and NSF Expedition `Mind in Vitro' award \#IIS-2123781. We also thank the computational support provided by the Bridges2 supercomputer at the Pittsburgh Supercomputing Center, through allocation TG-MCB190004 from the Extreme Science and Engineering Discovery Environment (XSEDE; NSF grant ACI-1548562).

\medskip

\bibliographystyle{MSP}
\bibliography{references.bib}

@inproceedings{Wang:2019,
	author = {Wang, Tixian and Taghvaei, Amirhossein and Mehta, Prashant G},
	booktitle = {2019 IEEE 58th Conference on Decision and Control (CDC)},
	organization = {IEEE},
	pages = {2758--2763},
	title = {Q-learning for POMDP: An application to learning locomotion gaits},
	year = {2019}}

@inproceedings{Zhang:2025,
	author = {Zhang, Chuhan and Wang, Cong and Pan, Wei and Della Santina, Cosimo},
	booktitle = {2025 IEEE 8th International Conference on Soft Robotics (RoboSoft)},
	organization = {IEEE},
	pages = {1--8},
	title = {SpikingSoft: A Spiking Neuron Controller for Bio-inspired Locomotion with Soft Snake Robots},
	year = {2025}}

@article{Naughton:2026,
	author = {Naughton, Noel and Tekinalp, Arman and Shivam, Keshav and Kim, Seung Hyun and Khairnar, Apoorva and Kindratenko, Volodymyr and Gazzola, Mattia},
	journal = {Proceedings of the National Academy of Sciences},
	number = {17},
	pages = {e2522094123},
	publisher = {National Academy of Sciences},
	title = {Neural reservoir control of a bio-hybrid soft arm},
	volume = {123},
	year = {2026}}

@article{Raffin:2021,
	author = {Raffin, Antonin and Hill, Ashley and Gleave, Adam and Kanervisto, Anssi and Ernestus, Maximilian and Dormann, Noah},
	journal = {Journal of machine learning research},
	number = {268},
	pages = {1--8},
	title = {Stable-baselines3: Reliable reinforcement learning implementations},
	volume = {22},
	year = {2021}}

@article{Bell-III:2022,
	author = {Bell III, James F and Maki, Justin N and Alwmark, Sanna and Ehlmann, Bethany L and Fagents, Sarah A and Grotzinger, John P and Gupta, Sanjeev and Hayes, Alexander and Herkenhoff, Ken E and Horgan, Briony HN and others},
	journal = {Science Advances},
	number = {47},
	pages = {eabo4856},
	publisher = {American Association for the Advancement of Science},
	title = {Geological, multispectral, and meteorological imaging results from the Mars 2020 Perseverance rover in Jezero crater},
	volume = {8},
	year = {2022}}

@article{Schulman:2017,
	author = {Schulman, John and Wolski, Filip and Dhariwal, Prafulla and Radford, Alec and Klimov, Oleg},
	journal = {arXiv preprint arXiv:1707.06347},
	title = {Proximal policy optimization algorithms},
	year = {2017}}

@article{Qi:2020,
	author = {Qi, Xinda and Shi, Hongyang and Pinto, Thassyo and Tan, Xiaobo},
	journal = {IEEE Robotics and Automation Letters},
	number = {2},
	pages = {1610--1617},
	publisher = {IEEE},
	title = {A novel pneumatic soft snake robot using traveling-wave locomotion in constrained environments},
	volume = {5},
	year = {2020}}

@article{Graham:2023,
	author = {Graham, Mal and Socha, John J},
	journal = {Journal of Experimental Biology},
	number = {19},
	pages = {jeb245094},
	publisher = {The Company of Biologists Ltd},
	title = {Dynamic gap crossing in Dendrelaphis, the sister taxon of flying snakes},
	volume = {226},
	year = {2023}}

@article{Tingle:2024,
	author = {Tingle, Jessica L and Garner, Kelsey L and Astley, Henry C},
	journal = {Annals of the New York Academy of Sciences},
	number = {1},
	pages = {16--37},
	publisher = {Wiley Online Library},
	title = {Functional diversity of snake locomotor behaviors: a review of the biological literature for bioinspiration},
	volume = {1533},
	year = {2024}}

@article{Jia:2021,
	author = {Jia, Yuanyuan and Ma, Shugen},
	journal = {IEEE Robotics and Automation Letters},
	number = {2},
	pages = {2319--2326},
	publisher = {IEEE},
	title = {A coach-based bayesian reinforcement learning method for snake robot control},
	volume = {6},
	year = {2021}}

@article{Bing:2022,
	author = {Bing, Zhenshan and Cheng, Long and Huang, Kai and Knoll, Alois},
	journal = {IEEE Robotics \& Automation Magazine},
	number = {4},
	pages = {92--103},
	publisher = {IEEE},
	title = {Simulation to real: learning energy-efficient slithering gaits for a snake-like robot},
	volume = {29},
	year = {2022}}

@article{Seetohul:2022,
	author = {Seetohul, Jenna and Shafiee, Mahmood},
	journal = {Robotics},
	number = {3},
	pages = {57},
	publisher = {MDPI},
	title = {Snake robots for surgical applications: A review},
	volume = {11},
	year = {2022}}

@article{Gimmel:2021,
	author = {Gimmel, Angela and {\"O}fner, Sabine and Liesegang, Annette},
	journal = {Journal of animal physiology and animal nutrition},
	pages = {24--28},
	publisher = {Wiley Online Library},
	title = {Body condition scoring (BCS) in corn snakes (Pantherophis guttatus) and comparison to pre-existing body condition index (BCI) for snakes},
	volume = {105},
	year = {2021}}

@article{Schiebel:2023,
	author = {Schiebel, Perrin E and Hubbard, Alex M and Goldman, Daniel I},
	journal = {Integrative and comparative biology},
	number = {1},
	pages = {198--208},
	publisher = {Oxford University Press},
	title = {Comparative study of snake lateral undulation kinematics in model heterogeneous terrain},
	volume = {63},
	year = {2023}}

@article{Fu:2022,
	author = {Fu, Qiyuan and Astley, Henry C and Li, Chen},
	journal = {Bioinspiration \& Biomimetics},
	number = {3},
	pages = {036009},
	publisher = {IOP Publishing},
	title = {Snakes combine vertical and lateral bending to traverse uneven terrain},
	volume = {17},
	year = {2022}}

@article{Kim:2023,
	author = {Kim, Yongdeok and Yang, Yiyuan and Zhang, Xiaotian and Li, Zhengwei and V{\'a}zquez-Guardado, Abraham and Park, Insu and Wang, Jiaojiao and Efimov, Andrew I and Dou, Zhi and Wang, Yue and others},
	journal = {Science Robotics},
	number = {74},
	pages = {eadd1053},
	publisher = {American Association for the Advancement of Science},
	title = {Remote control of muscle-driven miniature robots with battery-free wireless optoelectronics},
	volume = {8},
	year = {2023}}

@article{Shih:2023,
	author = {Shih, Chia-Hsien and Naughton, Noel and Halder, Udit and Chang, Heng-Sheng and Kim, Seung Hyun and Gillette, Rhanor and Mehta, Prashant G and Gazzola, Mattia},
	journal = {Advanced Intelligent Systems},
	number = {9},
	pages = {2300088},
	publisher = {Wiley Online Library},
	title = {Hierarchical control and learning of a foraging CyberOctopus},
	volume = {5},
	year = {2023}}

@article{Tekinalp:2024,
	author = {Tekinalp, Arman and Naughton, Noel and Kim, Seung Hyun and Halder, Udit and Gillette, Rhanor and Mehta, Prashant G and Kier, William and Gazzola, Mattia},
	journal = {Proceedings of the National Academy of Sciences},
	number = {41},
	pages = {e2318769121},
	publisher = {National Academy of Sciences},
	title = {Topology, dynamics, and control of a muscle-architected soft arm},
	volume = {121},
	year = {2024}}

@article{Zhang:2021,
	author = {Zhang, Xiaotian and Naughton, Noel and Parthasarathy, Tejaswin and Gazzola, Mattia},
	journal = {Nature communications},
	number = {1},
	pages = {1--8},
	publisher = {Nature Publishing Group},
	title = {Friction modulation in limbless, three-dimensional gaits and heterogeneous terrains},
	volume = {12},
	year = {2021}}

@article{Santina:2023,
	author = {{Della Santina}, Cosimo and Christian Duriez and Daniela Rus},
	doi = {10.1109/MCS.2023.3253419},
	issn = {1066-033X},
	journal = {IEEE Control Systems},
	language = {English},
	number = {3},
	pages = {30--65},
	publisher = {IEEE},
	title = {Model-Based Control of Soft Robots: A Survey of the State of the Art and Open Challenges},
	volume = {43},
	year = {2023}}

@article{Yasa:2023,
	author = {Yasa, Oncay and Toshimitsu, Yasunori and Michelis, Mike Y. and Jones, Lewis S. and Filippi, Miriam and Buchner, Thomas and Katzschmann, Robert K.},
	doi = {https://doi.org/10.1146/annurev-control-062322-100607},
	issn = {2573-5144},
	journal = {Annual Review of Control, Robotics, and Autonomous Systems},
	number = {Volume 6, 2023},
	pages = {1-29},
	publisher = {Annual Reviews},
	title = {An Overview of Soft Robotics},
	type = {Journal Article},
	url = {https://www.annualreviews.org/content/journals/10.1146/annurev-control-062322-100607},
	volume = {6},
	year = {2023}}

@article{Mengaldo:2022,
	author = {Gianmarco Mengaldo and Federico Renda and Brunton, {Steven L.} and Moritz B{\"a}cher and Marcello Calisti and Christian Duriez and Chirikjian, {Gregory S.} and Cecilia Laschi},
	doi = {10.1038/s42254-022-00481-z},
	issn = {2522-5820},
	journal = {Nature Reviews Physics},
	language = {British English},
	month = sep,
	number = {9},
	pages = {595--610},
	publisher = {Springer International Publishing},
	title = {A concise guide to modelling the physics of embodied intelligence in soft robotics},
	volume = {4},
	year = {2022}}

@article{Wang:2023,
	author = {Tianyu Wang and Christopher Pierce and Velin Kojouharov and Baxi Chong and Kelimar Diaz and Hang Lu and Daniel I. Goldman},
	doi = {10.1126/scirobotics.adi2243},
	eprint = {https://www.science.org/doi/pdf/10.1126/scirobotics.adi2243},
	journal = {Science Robotics},
	number = {85},
	pages = {eadi2243},
	title = {Mechanical intelligence simplifies control in terrestrial limbless locomotion},
	url = {https://www.science.org/doi/abs/10.1126/scirobotics.adi2243},
	volume = {8},
	year = {2023}}

@techreport{Ulrich:1988,
	author = {Ulrich, Nathan Thatcher},
	title = {Grasping with mechanical intelligence},
	year = {1988}}

@inproceedings{Tilton:2013,
	author = {Tilton, Adam K and Mehta, Prashant G and Meyn, Sean P},
	booktitle = {2013 American Control Conference},
	organization = {IEEE},
	pages = {2415--2421},
	title = {Multi-dimensional feedback particle filter for coupled oscillators},
	year = {2013}}

@inproceedings{Tilton:2014,
	author = {Tilton, Adam and Mehta, Prashant G},
	booktitle = {American Control Conference (ACC), 2014},
	organization = {IEEE},
	pages = {584--589},
	title = {Control with rhythms: A CPG architecture for pumping a swing},
	year = {2014}}

@inproceedings{Wang:2020,
	author = {Wang, Tixian and Taghvaei, Amirhossein and Mehta, Prashant G},
	booktitle = {2020 American Control Conference (ACC)},
	organization = {IEEE},
	pages = {2188--2193},
	title = {Bio-inspired learning of sensorimotor control for locomotion},
	year = {2020}}

@article{Yang:2013,
	author = {Yang, Tao and Mehta, Prashant G and Meyn, Sean P},
	journal = {IEEE transactions on Automatic control},
	number = {10},
	pages = {2465--2480},
	publisher = {IEEE},
	title = {Feedback particle filter},
	volume = {58},
	year = {2013}}

@article{Alben:2013,
	author = {Alben, S.},
	journal = {Proceedings of the Royal Society of London A},
	number = {2159},
	pages = {20130236},
	title = {Optimizing snake locomotion in the plane},
	volume = {469},
	year = {2013}}

@book{Antman:1973,
	author = {Antman, S.S.},
	booktitle = {Linear Theories of Elasticity and Thermoelasticity},
	pages = {641--703},
	publisher = {Springer},
	title = {The theory of rods},
	year = {1973}}

@article{Astley:2015,
	author = {Astley, Henry C and Gong, Chaohui and Dai, Jin and Travers, Matthew and Serrano, Miguel M and Vela, Patricio A and Choset, Howie and Mendelson, Joseph R and Hu, David L and Goldman, Daniel I},
	journal = {PNAS},
	number = {19},
	pages = {6200--6205},
	publisher = {National Acad Sciences},
	title = {Modulation of orthogonal body waves enables high maneuverability in sidewinding locomotion},
	volume = {112},
	year = {2015}}

@article{Aydin:2019,
	author = {Aydin, Onur and Zhang, Xiaotian and Nuethong, Sittinon and Pagan-Diaz, Gelson J and Bashir, Rashid and Gazzola, Mattia and Saif, M Taher A},
	journal = {PNAS},
	number = {40},
	pages = {19841--19847},
	publisher = {National Acad Sciences},
	title = {Neuromuscular actuation of biohybrid motile bots},
	volume = {116},
	year = {2019}}

@article{Gazzola:2018,
	author = {Gazzola, M. and Dudte, L. H. and McCormick, A. G. and Mahadevan, L.},
	journal = {Royal Society Open Science},
	number = {6},
	publisher = {The Royal Society},
	title = {Forward and inverse problems in the mechanics of soft filaments},
	volume = {5},
	year = {2018}}

@article{Goldman:2010,
	author = {Goldman, Daniel I and Hu, David L},
	journal = {American Scientist},
	number = {4},
	pages = {314--323},
	publisher = {JSTOR},
	title = {Wiggling Through the World: The mechanics of slithering locomotion depend on the surroundings},
	volume = {98},
	year = {2010}}

@article{Gray:1946,
	author = {Gray, J.},
	journal = {Journal of Experimental Biology},
	number = {2},
	pages = {101--120},
	title = {The mechanism of locomotion in snakes},
	volume = {23},
	year = {1946}}

@article{Gray:1950,
	author = {Gray, J. and Lissmann, H.W.},
	journal = {Journal of Experimental Biology},
	number = {4},
	pages = {354--367},
	title = {The kinetics of locomotion of the grass-snake},
	volume = {26},
	year = {1950}}

@book{Hirose:1993,
	author = {Hirose, S.},
	publisher = {Oxford University Press, Oxford},
	title = {Biologically Inspired Robots: Snake-Like Locomotors and Manipulators},
	year = {1993}}

@article{Hu:2009,
	author = {Hu, D.L. and Nirody, J. and Scott, T. and Shelley, M.J.},
	journal = {PNAS},
	number = {25},
	pages = {10081--10085},
	title = {The mechanics of slithering locomotion},
	volume = {106},
	year = {2009}}

@article{Jayne:1988,
	author = {Jayne, Bruce C},
	journal = {Journal of Morphology},
	number = {2},
	pages = {159--181},
	publisher = {Wiley Online Library},
	title = {Muscular mechanisms of snake locomotion: an electromyographic study of lateral undulation of the Florida banded water snake (\textit{Nerodia fasciata}) and the yellow rat snake (\textit{Elaphe obsoleta})},
	volume = {197},
	year = {1988}}

@article{Jayne:2020,
	author = {Jayne, Bruce C},
	journal = {Integrative and Comparative Biology},
	title = {What Defines Different Modes of Snake Locomotion?},
	year = {2020}}

@article{Marvi:2013,
	author = {Marvi, H. and Bridges, J. and Hu, D.L.},
	journal = {Journal of The Royal Society Interface},
	number = {84},
	pages = {20130188},
	title = {Snakes mimic earthworms: propulsion using rectilinear travelling waves},
	volume = {10},
	year = {2013}}

@article{Marvi:2014,
	author = {Marvi, H. and Gong, C. and Gravish, N. and Astley, H. and Travers, M. and Hatton, R.L.and Mendelson, J.R. and Choset, H. and Hu, D.L and Goldman, D.I.},
	journal = {Science},
	number = {6206},
	pages = {224--229},
	title = {Sidewinding with minimal slip: Snake and robot ascent of sandy slopes},
	volume = {346},
	year = {2014}}

@article{Mosauer:1932,
	author = {Mosauer, Walter},
	journal = {Science},
	number = {1982},
	pages = {583--585},
	publisher = {JSTOR},
	title = {On the locomotion of snakes},
	volume = {76},
	year = {1932}}

@article{Naughton:2021,
	author = {Naughton, Noel and Sun, Jiarui and Tekinalp, Arman and Parthasarathy, Tejaswin and Chowdhary, Girish and Gazzola, Mattia},
	journal = {IEEE Robotics and Automation Letters},
	number = {2},
	pages = {3389-3396},
	title = {Elastica: A Compliant Mechanics Environment for Soft Robotic Control},
	volume = {6},
	year = {2021}}

@article{Onal:2013,
	author = {Onal, C.D. and Rus, D.},
	journal = {Bioinspiration and biomimetics},
	number = {2},
	pages = {026003},
	title = {Autonomous undulatory serpentine locomotion utilizing body dynamics of a fluidic soft robot},
	volume = {8},
	year = {2013}}

@article{Pagan-Diaz:2018,
	author = {Pagan-Diaz, Gelson J and Zhang, Xiaotian and Grant, Lauren and Kim, Yongdeok and Aydin, Onur and Cvetkovic, Caroline and Ko, Eunkyung and Solomon, Emilia and Hollis, Jennifer and Kong, Hyunjoon and others},
	journal = {Advanced Functional Materials},
	pages = {1801145},
	publisher = {Wiley Online Library},
	title = {Simulation and Fabrication of Stronger, Larger, and Faster Walking Biohybrid Machines},
	year = {2018}}

@article{Pfeifer:2007,
	author = {Pfeifer, Rolf and Lungarella, Max and Iida, Fumiya},
	journal = {Science},
	number = {5853},
	pages = {1088-1093},
	title = {Self-Organization, embodiment, and biologically inspired robotics},
	volume = {318},
	year = {2007}}

@article{Rieser:2021,
	author = {Rieser, Jennifer M and Tingle, Jessica L and Goldman, Daniel I and Mendelson, Joseph R and others},
	journal = {PNAS},
	number = {6},
	publisher = {National Acad Sciences},
	title = {Functional consequences of convergently evolved microscopic skin features on snake locomotion},
	volume = {118},
	year = {2021}}

@article{Socha:2002,
	author = {Socha, John J},
	journal = {Nature},
	number = {6898},
	pages = {603--604},
	publisher = {Nature Publishing Group},
	title = {Gliding flight in the paradise tree snake},
	volume = {418},
	year = {2002}}

@inproceedings{Taghvaei:2014,
	author = {Taghvaei, Amirhossein and Hutchinson, Seth A and Mehta, Prashant G},
	booktitle = {53rd IEEE Conference on Decision and Control},
	organization = {IEEE},
	pages = {3487--3492},
	title = {A coupled oscillators-based control architecture for locomotory gaits},
	year = {2014}}

@article{Tanev:2005,
	author = {Tanev, I. and Ray, T. and Buller, A.},
	journal = {IEEE Transactions on Robotics},
	number = {4},
	pages = {632--645},
	title = {Automated evolutionary design, robustness, and adaptation of sidewinding locomotion of a simulated snake-like robot},
	volume = {21},
	year = {2005}}

@article{Transeth:2008,
	author = {Transeth, A.A. and Leine, R.I. and Glocker, C. and Pettersen, K.Y. and Liljeback, P.},
	journal = {IEEE Transactions on Robotics},
	number = {1},
	pages = {88--104},
	title = {Snake robot obstacle-aided locomotion: Modeling, simulations, and experiments},
	volume = {24},
	year = {2008}}

@article{Wang:2021,
	author = {Wang, J. and Zhang, X. and Park, J. and Park, I. and Kilicarslan, E. and Kim, Y. and Dou, Z. and Bashir, R. and Gazzola, M.},
	journal = {Advanced Intelligent Systems},
	pages = {2000237},
	title = {Computationally assisted design and selection of maneuverable biological walking machines},
	year = {2021}}

@article{Zhang:2019,
	author = {Zhang, Xiaotian and Chan, Fan Kiat and Parthasarathy, Tejaswin and Gazzola, Mattia},
	journal = {Nature communications},
	number = {1},
	pages = {1--12},
	publisher = {Nature Publishing Group},
	title = {Modeling and simulation of complex dynamic musculoskeletal architectures},
	volume = {10},
	year = {2019}}

\end{document}